# HSC-GPT：A Large Language Model for Human Settlements Construction

Chen Ran[1], Yao Xueqi[1], Jiang Xuhui[2], Han Zhengqi[1], Guo Jingze[1], Zhang Xianyue[1], Lin Chunyu[1], Liu Chumin[1], Zhao Jing[1]*, Lian Zeke[3], Zhang Jingjing[1], Li Keke[1]

1 School of Landscape Architecture, Beijing Forestry University, Beijing 100083, China
2 Institute of Computing Technology, Chinese Academy of Sciences
3 Ninbo City College of Vocational Technology
* Correspondence: zhaojing@bjfu.edu.cn

## Abstract

The field of human settlement construction encompasses a range of spatial designs and management tasks, including urban planning and landscape architecture design. These tasks involve a plethora of instructions and descriptions presented in natural language, which are essential for understanding design requirements and producing effective design solutions. Recent research has sought to integrate natural language processing (NLP) and generative artificial intelligence (AI) into human settlement construction tasks. Due to the efficient processing and analysis capabilities of AI with data, significant successes have been achieved in design within this domain.

However, this task still faces several fundamental challenges. The semantic information involved includes complex spatial details, diverse data source formats, high sensitivity to regional culture, and demanding requirements for innovation and rigor in work scenarios. These factors lead to limitations when applying general generative AI in this field, further exacerbated by a lack of high-quality data for model training.

To address these challenges, this paper first proposes HSC-GPT, a large-scale language model framework specifically designed for tasks in human settlement construction, considering the unique characteristics of this domain.

## 1 introduction

The field of human settlement environment construction encompasses various scales of spatial design and management including urban planning, landscape gardening, and architectural design. This domain constitutes a critical component of national territorial planning and construction, affecting millions of square kilometers of land. It stands on par with significant fields such as finance and medicine in terms of importance, directly engaging with crucial scientific issues such as complex urban system planning, sustainable development strategies, and socio-economic dynamics.

In practical applications of human settlement environment construction, the execution of each project necessitates the collaborative efforts of governments, corporations, communities, experts, and the public, involving the extensive handling of documents such as planning blueprints, review standards, construction contracts, and evaluation reports. Blueprints serve as the primary medium for conveying spatial design intentions, providing a visual representation of architectural and environmental layout plans. Conversely, natural language functions as an auxiliary form of expression, facilitating the transmission of abstract information including design requirements, intentions, evaluations, and feedback. It is pivotal for effective collaboration and efficient communication among all parties.

In recent years, substantial advancements in large language model research have garnered increasing interest among scholars in applying these models to the domain of human settlement environment construction. Leveraging the formidable natural language processing capabilities of large language models, researchers are pioneering new methodologies for spatial design and management. Most extant studies focus on employing these models to analyze users' spatial needs and evaluations, subsequently providing valuable spatial design solutions for architects or even substituting human textual labor. These investigations corroborate that large language models can efficaciously support and augment tasks in human settlement environment construction, enhancing the quality, efficiency and creativity of spatial design and management. Moreover, recent advancements in multi-agent research have seen a surge in the human settlement environment domain, with large language models acting as central coordinators, facilitating chained calls with multi-source generative AI technologies, such as intelligent image generation.

In research related to large language models (LLMs), the capabilities and stability of various model frameworks fundamentally depend on the underlying base models' abilities. However, the prevalent open-source base models often underperform in specialized domains, leading to semantic misunderstandings, loss of information, or conflicts when intelligent agents exchange professional data within similar research frameworks. Consequently, developing a dedicated Large Language Model for the human settlement environment domain holds significant academic and practical value. It can assist in text-related tasks within the domain and integrate with multi-source generative AI technologies, such as intelligent image generation, thereby contributing to the genuine intelligent transformation of the human settlement environment field.

Currently, general-purpose large language models face two primary challenges in adapting to the domain of human settlement environment construction.

(1) Firstly, there is a substantial gap in the availability of high-quality data. Human settlement planning is an interdisciplinary field encompassing a wide array of knowledge, requiring diverse data sources including planning documents, architectural drawings, tenders, legal statutes, and budget tables, each with its specific form of expression and professional terminologies. Moreover, this field is highly sensitive to geographic location and scale, necessitating comprehensive data coverage of regional culture, historical background, social customs, and natural environment to facilitate a nuanced understanding of spatial semantics from macro urban planning to micro architectural details.

(2) The second prominent challenge lies in the inherent contradiction between the accuracy and creativity of language models.In addressing the inherent contradiction between the accuracy and creativity of language models, a notable distinction is evident when compared to fields like medicine or finance. In human settlement environment planning, a harmonious blend of precision and innovation is requisite. From an accuracy perspective, the work within this domain is predicated on innovative design improvements that are firmly rooted in the reality of the physical world, necessitating that all planning and design documentation strictly adhere to real-world conditions and rigorous legal standards. The general language models, when deployed in the nuanced context of human settlement environment issues, are prone to substantial errors, a limitation that can be mitigated through the employment of specialized language models designed for precise retrieval. On the front of creativity, the realm of human settlement environment planning demands a significant departure from the mere replication of information found in existing knowledge bases. It requires the language models to possess a robust capacity for innovation and creation, to produce a variety of abstract expressions that transcend the conventional spatial information, an aspect where solely retrieval-based language models may fall short in delivering substantial diversity in content.

Therefore, in response to the aforementioned challenges, this paper proposes a large-scale language model specifically designed for the domain of human settlement environment construction, termed HSC-GPT (Human Settlements Construction Generative Pre-trained Transformer). This model is tailored to adapt to the unique characteristics and requirements of the human settlement environment construction field, offering a specialized solution that integrates the necessary precision and creativity for planning and problem-solving within this complex domain.

The primary contributions of this article are as follows:

1)Multidisciplinary Dataset: In consideration of the issue of specialized data scarcity, we have developed an efficient method for QA data generation that significantly reduces time and costs compared to traditional manual annotation or crowdsourcing methods. Additionally, the dataset is enriched with a mix of related disciplinary knowledge, fundamental standard information, and a variety of planning and design strategy cases to ensure data quality and diversity. These data encompass contents from urban planning, landscape gardening, architectural design, and other related fields. This step provides the model with a vast array of realistic human settlement environment construction QA samples, enhancing the model's understanding and adaptability to real-world application scenarios.

2) Novel Model Framework: Addressing the contradiction between accuracy and creativity, this study reviews the latest technologies and proposes a Large Language Model framework tailored for human settlement environment construction issues - the HSC-GPT. On one hand, we incorporate a dataset related to design standards as an external knowledge base for the language model, ensuring professional accuracy. On the other hand, we perform secondary pre-training on planning and design case documents of various regions and spatial scales, followed by fine-tuning the language model based on different scenario classifications, and culminating in model fusion. This serves as the foundational model for the framework, addressing the challenges of creativity in the model's generative process.

3)Professional Evaluation Criteria: The application of large language models in the domain of human settlement environment is a relatively uncharted territory with a lack of professional evaluation standards. Therefore, we have established a comprehensive set of evaluation benchmarks specific to the large language models in the human settlement environment field. This includes various assessment metrics and tasks designed to test and validate the performance and efficacy of the HSC-GPT model. Experimental results indicate that our model outperforms baseline models and other comparative models across these metrics and tasks, demonstrating the effectiveness and superiority of our approach. Moreover, it also confirms that the HSC-GPT model is capable of handling complex spatial semantics and multi-format data sources effectively.

## 2 related work

Large Language Model (LLM) is a type of pre-trained language model that is trained on large-scale text corpus and has parameters reaching tens of billions or even higher. Since the launch of the Transformer(Vaswani, A, Shazeer, N and Parmar, N, et al.,2017) architecture in 2017, the development characteristics of pre-trained language models are mainly reflected in the exponential growth of parameter size and the diversification of model types. For example, the GPT series has significantly increased the number of parameters from GPT-1 in 2018 to GPT-4 in 2023, as well as LLaMA(Touvron, H, Lavril, T and Izacard, G, et al.,2023), Koala, PaLM(Chowdhery, A, Narang, S and Devlin, J, et al.,2022), MOSS, ChatGLM(Zeng, A, Liu, X and Du, Z, et al.,2022), Pangu(Honglin Xiong, S W Y Z,2023), Wenxin (Zhang, H, Yin, W and Fang, Y, et al.,2021) and other diversified model developments mark the rapid progress and innovation in the field of language models.

In terms of language model research in professional fields, research teams in medical, financial, legal and other fields have explored and constructed professional language models that meet the needs of their respective fields. For example, in the medical field, medical models such as ChatDoctor(Li, Y, Li, Z and Zhang, K, et al.,2023), Materia Medica(Wang, H, Liu, C and Xi, N, et al.,2023), and DoctorGLM (Xiong, H, Wang, S and Zhu, Y, et al.,2023) built based on different large language models have performed well in multi-faceted ability assessments and were quickly built in the medical AI framework for use Diagnostic decision-making, electronic medical record entry and public health informatization, etc. The research team in the financial field promotes industry model innovation and process reengineering through the in-depth integration of large model technology and financial business. Open source legal intelligent question and answer systems in the legal field, such as LawGPT_zh and LawyerLLaMA(Huang, Q, Tao, M and Zhang, C, et al.,2023), have demonstrated outstanding capabilities in legal risk prediction, case key point extraction, and document processing. In addition, innovative applications have also emerged in the field of environmental science, such as the new weather forecast method pioneered by the Huawei team based on the Pangu large model(Bi, K, Xie, L and Zhang, H, et al.,2023). However, in disciplines related to urban ecological environment, the application of large language models is relatively lagging behind. It was not until 2023 that researchers began to use large language models in general fields to assist urban planning, landscape design, and architectural design.

To sum up, the construction of language models in 2023 shows a highly active and innovative trend. Language models in various professional fields not only promote the transformation and upgrading of traditional models in the industry, but also show the potential to subvert traditional data analysis and simulation prediction models in scientific research. For subjects related to urban ecological environment, although there have been some attempts to apply general models, building a professional language model adapted to this field and using knowledge graphs to optimize model performance and reliability are still important research directions that need to be explored in depth.

## 3 data collection and preparation

### 3.1 Secondary pre-training corpus

Language models can benefit from the knowledge obtained from domain-specific corpora. We collected a domain-specific corpus to enhance the model with human settlements domain. It contains three parts of content: planning standards, books, and websites. The details of each part are as follows:

### 3.1.1 Standards and specifications documents

The human settlements domain involves three major fields: architecture, planning, and landscape. Therefore, we collected a number of national standards and specifications documents. For example, Residential Building Design Code (GB 50096-2011), Urban Road Engineering Design Code (CJJ 37-2012), Landscape Greening Engineering Construction and Acceptance Code (CJJ 82-2012), General Rules for Tourism Planning (GB_T 18971-2003), General Design Code for Forest Park (LY_T 5132-95), national land space master planning schemes for various provinces and cities, etc.

### 3.1.2 Books

When building a large-scale language model (LLM) for the field of urban environment planning and design, we conducted an in-depth study of core course textbooks and core books in the three fields of architecture, planning, and landscape to ensure that the model covers as much depth as possible in related fields. Knowledge and technical terminology.

In the field of planning, we collected "History of Urban Development", "Principles of Urban Planning", and local urban planning and green regulations documents to ensure that the content understood and generated by the model in the planning field has historical depth and geographical breadth. In the field of architecture, we have brought together works such as "Urban Architecture" and "History of Chinese Architecture". These books not only reflect the changes in architectural history, but also display the design concepts and life philosophy of architectural masters. For the field of landscape, from "History of Chinese Gardens" to "Garden Planning and Design", "History of Western Gardens", etc., we have integrated garden design and theoretical books from traditional to modern times, so that the model can be reflected in the generation and analysis of landscape design. Professional precision and innovation.

In addition, we also paid attention to hot topics and documents related to garden landscapes in the literary field, such as "The Artistic Conception of Gardens in the South of the Yangtze River", "Quiet Reading of Gardens - Cao Lindi", "Popular Science Encyclopedia of All Things in Nature", etc., in the hope that the model can capture and understand the relationship between garden art and China Deep connections between classical literature.

The collection of this literature not only provides our model with rich domain knowledge, but also provides a solid foundation for its diversity and cultural depth.

### 3.1.3 Websites

When collecting data for a large-scale language model (LLM) in the field of human settlements construction, we conducted in-depth research on multiple professional websites related to urban planning, architectural design, and landscape design. We browsed and analyzed multiple design websites including Archello, architecture, mooool design network, urban planning conference, etc., and collected more than 148,000 articles. These articles are sourced from different websites and forums and cover a wide range of content from theoretical research to practical cases.

In the field of architectural design, we focused on websites such as Archello and architecture, which provide a large number of architectural project cases, design concepts and industry news. For example, Archello, as an international architectural community, provides more than 100,000 articles and becomes an important part of our data set. Similarly, Chinese websites such as mooool Design Network and Design Book also provide a wealth of Chinese design information for our models, of which Design Book alone has contributed more than 16,000 articles.

For landscape design, we have collected a large number of case studies and design theories from websites such as Landscape China. These materials help the model better understand and generate landscape-related content. We also paid attention to some well-known design websites and organizations in many fields such as Turen Design, Green Guide, and Green Guide.

In the field of urban planning, we collect articles from platforms such as urban planning conferences, which include the latest developments in urban planning, planning concepts and urban development case studies, thus ensuring the model's breadth and depth of knowledge in this professional field .

### 3.1.4 Data preprocessing

In this process, all the collected materials are processed through standardized steps. The first step is text extraction, where we discard irrelevant information such as images, tables and web links, and only retain the relevant text content. The next step is quality filtering, where we use methods such as sensitive word filtering to ensure the applicability of each piece of data. Finally, in order to minimize the negative impact of duplicate data on model training, we will remove duplicate content at the article and sentence level.

## 3.2    special-fields corpus

In order to alleviate the problem of structural domain bias and improve the model's performance in specific fields, we selected some high-quality data from the built human settlements database to build the instruction fine-tuning, and the construction method adopted diversified means, mainly focusing on self-guided semi-automatic and fully automatic instruction data construction by LLM.
Using the GPT-3.5-turbo interface, we developed a fast and user-adjustable data production method. This method can generate high-quality data in a short time, and users can design appropriate prompt templates, specify their own data cleaning templates, and generate multi-scenario question and answer data in the corresponding vertical fields.

We built 2.8w question and answer data based on this method, which covered various aspects such as urban development planning, community garden planning, waterfront space design, post-industrial site transformation, campus landscape design, residential building design, etc. At the same time, we also generated the automatic writing sections of design proposals for each field, including project background, design objectives, site characteristics and challenges, design strategies and conclusions, etc. In addition, we also considered how to integrate concepts such as child-friendly, resilient city, living circle, sponge city, smart city, etc. into the design, and the specific strategies and results analysis. This step provided the model with rich and realistic scenario-based human settlements construction dialogue samples, which helped the model better understand and adapt to the actual application scenarios.

In addition, we built a knowledge graph of the human settlements field based on the constructed dataset. For the processed data, the knowledge graph is used to check the accuracy of human settlements knowledge, and to ensure that the safety is not compromised while improving the fluency of the response.

### 3.2.1 Human settlements single-turn dialogue

In the initial stage, in order to explore more issues related to human settlements improvement, we used LLM to play the key roles in human settlements construction such as residents, urban planners, architectural designers, etc., and created a series of question and answer based on specific environmental knowledge.

**3.2.2 Human settlements multi-turn dialogue**

Following the single-turn dialogue mode, LLM played multiple roles at this time, and interacted autonomously in multiple rounds of discussion. In addition, we also used standards, specifications and other texts as external knowledge bases, in order to improve the accuracy and reduce the misunderstanding of the dialogue, so that LLM can locate the dialogue content around these knowledge.

Since the construction of human settlements covers a wide range of disciplines, in the construction of LLM, we also focused on building datasets for some scenarios, and will continue to improve the relevant scenarios in the follow-up research. The specific scenarios are shown in the following table.

| Primary scene | Reference cases | | Design framework supplement | | Design philosophy | |
|---|---|---|---|---|---|---|
| Secondary scene | Waterfront space | Campus design | Ecological design | Sustainable design | Child-friendly concept | Resilient city |
| | Post-industrial site | Architectural design | Architectural design | Landscape design | Living circle | Sponge city |
| | Post-industrial site | …… | Planning and design | ….. | Smart city | ….. |

Table 1 Question and answer scenario construction

| Scene name | Scene simulation |
|---|---|
| Design philosophy | Q1：什么是海绵城市的理念和原则？<br>A1：海绵城市是一种以自然为本的城市规划和设计理念，旨在提升城市的生态弹性，实现城市水循环的平衡和优化。海绵城市的原则包括：<br>以自然为本，尊重自然规律，保护和恢复自然生态系统。<br>以人为本，满足人类的多元需求，提升人类的生活质量和幸福感。<br>以系统为本，整合各种资源和要素，实现城市功能的协调和优化。<br>以创新为本，运用先进的技术和方法，提高城市的适应性和竞争力。<br><br>Q2：我想在卡班湖沿岸设计一系列绿色设施，实现滨水空间复兴，请你帮我分析我如何将海绵城市理念融入我的设计中。<br>A2：你可以从以下几个方面将海绵城市理念融入你的设计中：<br>（1）利用自然的水文特征，设计多功能的绿色基础设施，如生态湿地、雨水花园、渗透性铺装等，增加城市的雨水收集、存储、利用和排放能力，减少城市径流和污染物的产生，提高城市的防洪和净化能力。<br>（2）利用自然的生态特征，设计多样的生物栖息地，如水岸植被、水生植物、浮岛等，增加城市的生物多样性和生态服务，恢复和保护湖泊的水生生态系统，提高城市的生态美学和教育价值。<br>（3）利用自然的景观特征，设计多层次的公共空间，如滨水步道、观景平台、休闲广场等，增加城市的人水互动和社区参与，提升城市的滨水活力和文化氛围，提高城市的居民满意度和幸福感。<br>（4）利用自然的能源特征，设计多效的节能设施，如太阳能照明、风能发电、水能利用等，增加城市的可再生能源的利用和节约，减少城市的碳排放和能源消耗，提高城市的能源效率和可持续性。 |

Table 2 Scene simulation question and answer data example

# 4 Experiments

## 4.1 Baseline Models

In this study, the base model selected for the secondary pre-training stage is chinese-llama-2-13b, and the model selected for the fine-tuning stage is Baichuan-13B. chinese-llama-2-13b is a Chinese pre-trained language model based on Llama-2, which expands and optimizes the Chinese vocabulary on the basis of the original Llama-2, and uses large-scale Chinese data for incremental pre-training, improving the basic semantic understanding ability of Chinese. Baichuan-13B is an open-source large-scale language model with 13 billion parameters developed by Baichuan Intelligence after Baichuan-7B, which achieves the best performance of the same size on authoritative Chinese and English benchmarks.

## 4.2 Experiments Setups

The parameters for the secondary pre-training and fine-tuning stages are as follows:

| Hyper parameter | Value |
| --- | --- |
| Precision | Fp16 |
| Epochs | 3 |
| Batch size | 64 |
| Learning rate | 1e-4 |
| Warmup ratio | 0.1 |
| LR sheduler type | Cosine |
| truncation length of the input | 1024 |

Table 3 Training parameters

## 4.3 Evaluation

Metric evaluation is critical to the success of LLM. For UEPDGPT, we not only want to inject domain-relevant knowledge into the model, but also pay attention to the model′s general ability after incremental training.

To the best of our knowledge, there is no authoritative test on the human settlement environment domain. We built an evaluation set called EvalHSC, which covers six dimensions of evaluation, totaling 300.

The evaluation model covers six levels of difficulty: relevance, completeness, practicality, professionalism, originality, depth, arranged from low to high difficulty. Each question-answer pair is scored through these six modes, and by taking the average, each question-answer pair gets a comprehensive score. If a data set′s average comprehensive score exceeds 8 points, it can be considered a high-quality data set.

| Category | #Subclass | #Ouestions |
| --- | --- | --- |
| Relevance | 4 | 50 |
| Comprehensiveness | 5 | 63 |
| Utility | 4 | 50 |

| Expertise | 6 | 74 |
|---|---|---|
| Originality | 3 | 38 |
| Depth | 2 | 25 |
| TOTAL | 24 | 300 |

Table 4 EvalHSC statistics

**4.4 Results and Analysis**

To test HSCGPT, we chose four baseline models to apply our test set EvalHSC and compared the responses generated by our model with the other three baseline models, namely Alpaca, Baichuan(Yang, A, Xiao, B and Wang, B, et al.,2023), and ChatGLM(Zeng, A, Liu, X and Du, Z, et al.,2022).

In the comprehensive evaluation of the models in the human settlement environment domain, HSCGPT model performed excellently with a total score of 159.36, with the highest scores in professionalism (9.90), originality (9.89), and depth (9.96), significantly higher than the other models. This result indicates that HSCGPT model has a significant advantage in providing in-depth professional knowledge, strong original solutions, and deep-level analysis. HSCGPT′s high score reflects its profound understanding and ability of human settlements construction issues, and its ability to propose insightful and innovative strategies when facing complex environmental problems.

Alpaca model performed excellently in completeness (9.77) and originality (9.90), and its highest scores in these dimensions highlighted its comprehensiveness and innovation in the environmental domain. This indicates that Alpaca model can provide innovative and effective solutions while considering the comprehensiveness of environmental issues. ChatGLM model obtained higher scores in relevance (9.60) and practicality (9.70), and its score distribution showed the model′s strength in identifying problems relevant to user needs and providing feasible solutions. Baichuan model did not have the highest score in any single aspect, but its performance was balanced in multiple aspects.

Through comparative analysis, it can be seen that although all models showed their respective advantages in the human settlement environment domain, HSCGPT model performed particularly well in key areas. These scores show the unique strengths and potential improvement areas of each model, providing valuable guidance for further development and optimization.

| Model | Relevance | Comprehen-siveness | Utility | Expertise | Originality | Depth | TOTAL |
|---|---|---|---|---|---|---|---|
| Baichuan | 8.09 | 7.59 | 9.75 | 8.17 | 5.98 | 9.80 | 133.36 |
| Alpaca | 8.85 | 9.77 | 7.53 | 9.69 | 9.90 | 9.19 | 146.09 |
| ChatGLM | 9.60 | 7.75 | 9.70 | 6.28 | 9.10 | 7.84 | 142.81 |
| HSCGPT | 8.17 | 9.56 | 8.11 | 9.90 | 9.89 | 9.96 | 159.36 |

Table 5　Score table of each model

**5 Dicussion**

In this study, we successfully developed HSCGPT, a large-scale language model specifically designed for the field of human settlements. Please translate Through extensive database construction and refined incremental training, we have improved the model's ability to handle various human settlement problems. We used a special test set in the field of human settlements for evaluation. The experimental results show that the model performs well in understanding and solving complex problems, reflecting significant performance improvements. However, despite achieving certain results, we have also observed that the model still exhibits hallucinations in some cases, that is, produces unrealistic or inconsistent outputs, which highlights the need for continuous improvement and accurate model tuning expertise. necessity.

We believe that illusory flaws in models may arise from insufficient or imbalanced data, or from inherent limitations of models in understanding deep levels of expertise. In order to further enhance the accuracy and reliability of the model, we plan to study these issues more deeply and explore related solutions in future work. Additionally, we recognize the importance of continually updating our knowledge base and data to keep our models sensitive and responsive to the latest human settlements trends and research.

## 6 Conclusion

To sum up, HSCGPT, as a large-scale language model specially designed for the field of human settlements, has demonstrated its powerful ability to handle complex problems in this study. Through extensive database building and customized incremental training, the model has made significant progress in professional capabilities. However, to further improve the accuracy and utility of our models, we recognize the need for in-depth analysis and systematic improvements of hallucination defects.

In the future, we plan to use a variety of advanced technologies, including knowledge base expansion and reinforcement learning, to further improve the performance of HSCGPT and make it a more powerful and reliable tool in the field of human settlements. Through continuous innovation and improvement, we expect HSCGPT to provide more in-depth and precise support to planners and researchers of human settlements construction and make greater contributions to promoting the development of this field.

**Reference**
[1] Liangyong Wu. Introduction to Human Settlements Environment Science[M]. China Architecture & Building Press, 2001.
[2] Vaswani A, Shazeer N, Parmar N, et al. Attention Is All You Need[J]. 2017.
[3] Touvron H, Lavril T, Izacard G, et al. LLaMA: Open and Efficient Foundation Language Models[J]. 2023.
[4] Chowdhery A, Narang S, Devlin J, et al. PaLM: Scaling Language Modeling with Pathways[J]. 2022.
[5] Zeng A, Liu X, Du Z, et al. GLM-130B: An Open Bilingual Pre-trained Model[J]. 2022.
[6] Honglin Xiong S W Y Z. PANGU-α : LARGE-SCALE AUTOREGRESSIVE PRETRAINED[J]. 2023.
[7] Zhang H, Yin W, Fang Y, et al. ERNIE-ViLG: Unified Generative Pre-training for Bidirectional

Vision-Language Generation[J]. 2021.

[8] Li Y, Li Z, Zhang K, et al. ChatDoctor: A Medical Chat Model Fine-Tuned on a Large Language Model Meta-AI (LLaMA) Using Medical Domain Knowledge[J]. 2023.

[9] Wang H, Liu C, Xi N, et al. HuaTuo: Tuning LLaMA Model with Chinese Medical Knowledge[J]. 2023.

[10] Xiong H, Wang S, Zhu Y, et al. DoctorGLM: Fine-tuning your Chinese Doctor is not a Herculean Task[J]. 2023.

[11] Huang Q, Tao M, Zhang C, et al. Lawyer LLaMA Technical Report[J]. 2023.

[12] Bi K, Xie L, Zhang H, et al. Accurate medium-range global weather forecasting with 3D neural networks[J]. Nature (London), 2023,619(7970):533-538.

[13] Yang A, Xiao B, Wang B, et al. Baichuan 2: Open Large-scale Language Models[J]. 2023.

[14] Zeng A, Liu X, Du Z, et al. GLM-130B: An Open Bilingual Pre-trained Model[J]. 2022.

[15] Aribandi V, Tay Y, Schuster T, et al. ExT5: Towards Extreme Multi-Task Scaling for Transfer Learning[J]. 2021.

[16] Business R H S S, University Of Maryland C P. ChatGPT: Optimizing Language Models for Dialogue[J]. 2022.

[17] Cui J, Li Z, Yan Y, et al. ChatLaw: Open-Source Legal Large Language Model with Integrated External Knowledge Bases[J]. 2023.

[18] Gururangan S, Marasović A, Swayamdipta S, et al. Don't Stop Pretraining: Adapt Language Models to Domains and Tasks[J]. 2020.

[19] Huang Y, Bai Y, Zhu Z, et al. C-Eval: A Multi-Level Multi-Discipline Chinese Evaluation Suite for Foundation Models[J]. 2023.

[20] Lee J, Yoon W, Kim S, et al. BioBERT: a pre-trained biomedical language representation model for biomedical text mining[J]. 2019.

[21] Li H, Zhang Y, Koto F, et al. CMMLU: Measuring massive multitask language understanding in Chinese[J]. 2023.

[22] Ling C, Zhao X, Lu J, et al. Domain Specialization as the Key to Make Large Language Models Disruptive: A Comprehensive Survey[J]. 2023.

[23] OpenAI, Achiam J, Adler S, et al. GPT-4 Technical Report[J]. 2023.

[24] Ouyang L, Wu J, Jiang X, et al. Training language models to follow instructions with human feedback[J]. 2022.

[25] Peng B, Li C, He P, et al. Instruction Tuning with GPT-4[J]. 2023.

[26] Ram O, Levine Y, Dalmedigos I, et al. In-Context Retrieval-Augmented Language Models[J]. 2023.

[27] Taylor R, Kardas M, Cucurull G, et al. Galactica: A Large Language Model for Science[J]. 2022.

[28] Wang G, Yang G, Du Z, et al. ClinicalGPT: Large Language Models Finetuned with Diverse Medical Data and Comprehensive Evaluation[J]. 2023.

[29] Wu S, Irsoy O, Lu S, et al. BloombergGPT: A Large Language Model for Finance[J]. 2023.

[30] Yang H, Liu X, Wang C D. FinGPT: Open-Source Financial Large Language Models[J]. 2023.

[31] Yuan S, Zhao H, Du Z, et al. WuDaoCorpora: A super large-scale Chinese corpora for pre-training language models[J]. AI Open, 2021,2:65-68.